\documentclass[10pt,conference]{IEEEtran}
\usepackage{amsmath,amsfonts}
\usepackage{algorithmic}
\usepackage{algorithm}
\usepackage{array}
\usepackage[caption=false,font=normalsize,labelfont=sf,textfont=sf]{subfig}
\usepackage{textcomp}
\usepackage{stfloats}
\usepackage{url}
\usepackage{verbatim}
\usepackage{graphicx}
\usepackage{cite}
\usepackage{amsthm}
\usepackage{booktabs}
\usepackage{multirow}
\usepackage{xcolor}
\usepackage{hyperref}

\begin{document}

\title{MetaRTL: Meta-path Attention Enhanced Relational Table Learning}



\author{
\IEEEauthorblockN{
Ken Zhong\textsuperscript{1},
Weichen Li\textsuperscript{1},
Zheng Wang\textsuperscript{1,2*}
}

\IEEEauthorblockA{\textsuperscript{1}School of Computer Science, Shanghai Jiao Tong University, Shanghai, China}

\IEEEauthorblockA{\textsuperscript{2}Shanghai Key Laboratory of Integrated Administration Technologies for Information Security, Shanghai, China}

\IEEEauthorblockA{\{zhongken, weichenli, wzheng\}@sjtu.edu.cn}

\thanks{\textsuperscript{*}Corresponding author.}
}

\maketitle

\begin{abstract}
Relational table learning has gained increasing attention with the widespread use of relational databases. Existing methods typically rely on deep GNN or HGNN stacks, leading to high computational costs and limited performance on large real-world databases. We propose MetaRTL, a two-stage framework for scalable and expressive relational table learning. In the first stage, MetaRTL obtains initial table embeddings via lightweight pre-training. In the second stage, it performs non-parametric message passing to derive meta-path features, which are then aggregated by an attention module, MetaAttn. By shifting computation from deep message passing to efficient meta-path aggregation, MetaRTL captures rich relational semantics while maintaining high efficiency. Experiments on 10 real-world datasets across 24 tasks demonstrate the effectiveness of the proposed method.
\end{abstract}


\section{Introduction}

Relational databases serve as the backbone of data management in modern applications, supporting critical domains such as finance, healthcare, and e-commerce~\cite{halpin2010information}.
Consequently, machine learning on relational table data, referred to as relational table learning (RTL)~\cite{rllm}, has attracted research attention~\cite{Džeroski2010}.
In the era of deep learning, the goal of this task is to model relational table data in an end-to-end fashion, eliminating the need for labor-intensive feature engineering and achieving strong performance in various applications such as user profiling, behavior prediction, and click-through rate estimation~\cite{zahradnik2023deep}~\cite{chen2026tlsql}.

A common practice among existing deep RTL methods is first to construct heterogeneous graphs based on primary–foreign key (pkey–fkey) relationships~\cite{rdl}, where each node corresponds to a single row from a table and inter-table dependencies are captured via diverse edge types. Then, to process such structured relational data, these methods~\cite{rdl}~\cite{rllm} often adopt a direct yet practical approach that builds on advances in tabular neural networks (TNNs)~\cite{borisov2022deep} and graph neural networks (GNNs)~\cite{gcn} or heterogeneous GNNs (HGNNs)~\cite{zou2022exploring}. Typically, TNNs straightforwardly encode per-table semantics by generating initial node embeddings for each row, which are then simply refined through stacked GNN/HGNN layers to facilitate cross-table message passing.

While this paradigm has achieved some advanced results, it still faces at least two notable challenges in practical applications.
On one hand, the resulting heterogeneous graphs often exhibit high relational complexity and dense connectivity. For example, the publicly available Stack Exchange dataset~\cite{stackexchange}, which originates from a real-world question answering platform, contains eight relational tables per domain. These include entities such as users, posts, and comments, with each table containing on average tens of millions of records. 
On the other hand, existing methods rely on deep GNN/HGNN stacks, which may incur substantial computational costs and suffer from common deep GNN issues~\cite{li2018deeper}~\cite{wang2024cluster}.

To address these challenges, we propose \textbf{MetaRTL} (\underline{Meta}-path based \underline{R}elational \underline{T}able \underline{L}earning), a novel framework designed to enhance the modeling capacity of RTL on large-scale relational tables.
The core idea of MetaRTL is twofold: (1) it leverages meta-paths to capture rich semantic relationships within complex relational structures, and (2) it employs an attention mechanism to enable feature fusion, by emphasizing task-relevant meta-path features.
For ease of implementation and efficiency, MetaRTL operates in two stages:
\begin{enumerate}
\item \textbf{Pre-training Stage}: A table encoder, consisting of a set of TNNs, is trained with substantially fewer iterations than previous RTL frameworks. Instead of relying on deep GNN/HGNN stacks, we adopt a simple two-layer HGNN and terminate training early (5 epochs in our experiments), producing stable node embeddings sufficient for subsequent meta-path aggregation.
\item \textbf{Aggregation Stage}: These embeddings are then used to compute semantic features along meta-paths on the heterogeneous graph via non-parametric propagation. To effectively integrate these features while mitigating sampling-induced semantic bias, we propose \textbf{MetaAttn}, a lightweight attention module that combines meta-path self-attention with global node cross-attention.
\end{enumerate}

Intuitively, the initial embeddings from the Pre-training Stage provide a strong foundation, while the attention mechanism in the Aggregation Stage enables expressive and interpretable feature fusion.
From a computational perspective, by shifting the modeling focus to the aggregation stage and leveraging precomputed meta-path features, MetaRTL avoids deep GNN stacking pipelines while maintaining efficiency. 
As a result, it scales linearly with the number of table rows and relations, effectively handling complex relational table data.
Extensive experiments on ten real-world datasets covering 24 tasks demonstrate the effectiveness of the proposed method.


Our main contributions are summarized as follows:
\begin{itemize}
    \item We propose MetaRTL, a novel two-stage RTL framework based on meta-paths, capturing complex relational table information.
    \item We design a lightweight and efficient aggregation module, MetaAttn, which integrates meta-path features with global node semantics.
    \item We conduct extensive experiments on various real-world datasets, demonstrating the superiority of MetaRTL.
\end{itemize}


\begin{figure*}[!t]
    \centering
    \includegraphics[width=1\linewidth]{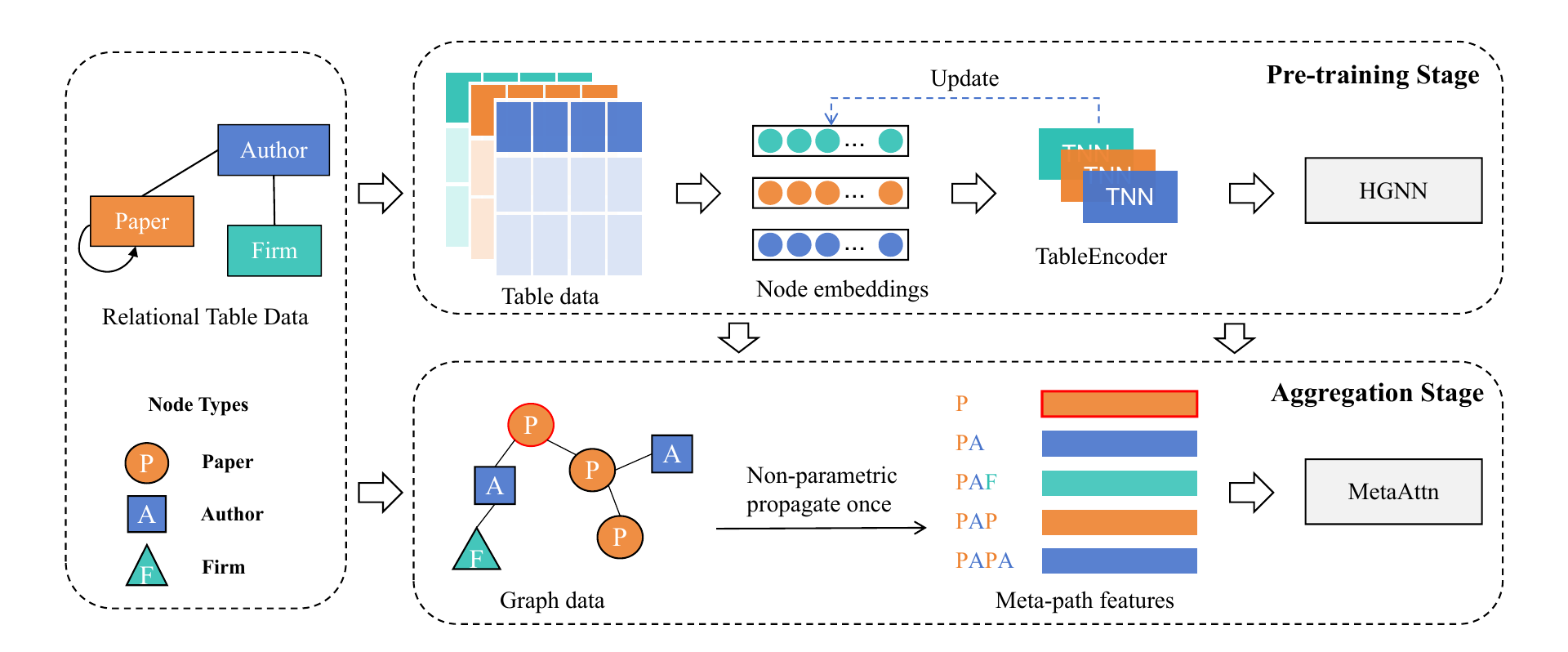}
    \caption{
    Overview of the \textbf{MetaRTL} framework. Given a relational table data, we first transform them into a heterogeneous graph following Definition~1 as a preprocessing step. The framework then proceeds in two stages. In the Pre-training stage (top), a unified table encoder composed of multiple TNNs is trained jointly with a shallow HGNN for very few epochs on neighbor sampled subgraphs, producing stable and semantically meaningful node embeddings. In the Aggregation stage (bottom), these embeddings are used to compute meta-path features through non-parametric propagation on the graph. A lightweight attention module, MetaAttn, then aggregates the multiple meta-path features and global node semantics to obtain task-specific target node representations.
    }
    \label{fig:1}
\end{figure*}
\section{Related Work}

\subsection{Heterogeneous Graph Neural Networks}
Heterogeneous Graph Neural Networks (HGNNs), a branch of Graph Neural Networks (GNNs) research~\cite{zou2022exploring}, are widely used for modeling multi-typed graph data.
Existing methods can be broadly classified as parametric or non-parametric based on whether they rely on trainable parameters.
Parametric HGNNs (like HAN~\cite{han}) employ trainable parameters to model semantic heterogeneity, by stacking architectures to aggregate multi-hop neighborhood information and compute attention weights during training.
However, these models are expensive to train on large graphs and often suffer from over-smoothing~\cite{li2018deeper}~\cite{wang2024cluster}~\cite{wang2021zero}.
Non-parametric HGNNs (like SeHGNN~\cite{sehgnn}) reduce the reliance on type-specific parameters by precomputing meta-path features before training.
This improves efficiency, but such methods usually depend on fixed node features and are not compatible with sampling-based mini-batch training.
MetaRTL builds upon this research direction and bridges the gap in applying such methods to relational table data, where node features are automatically learned from raw table entries and graph structures are induced by schema-level relations.

\subsection{Relational Table Learning}
Relational table learning (RTL) focuses on modeling structured data from real-world multi-table relational databases linked by primary–foreign key (pkey–fkey) relationships~\cite{Džeroski2010}. 
Early approaches explicitly modeled relational dependencies across tables using logic~\cite{kersting2006inductive} or probabilistic graphical models~\cite{getoor2007probabilistic}. While expressive, these methods often require manual feature engineering and do not scale well to large databases.
With the rise of deep learning, recent studies leverage techniques such as tabular neural networks (TNNs) and graph neural networks (GNNs)~\cite{wang2016database, zahradnik2023deep}
For example, graph-based representations across tables have been explored~\cite{Cvitkovic_2020}. 
RDL~\cite{rdl} introduces relational entity graphs and a universal TNN–GNN framework that generalizes to arbitrary multi-relational tables. BRIDGE~\cite{rllm} simplifies this approach by reducing the number of auxiliary tables and ignoring heterogeneity.
RelGNN~\cite{relgnn} further decomposes message passing into atomic path modeling based on relational insights. LightRDL~\cite{lightrdl} focuses on efficiency and distills relational patterns with small GNNs while relying on engineered features for temporal signals, enabling faster inference. 
MetaRTL differs from these approaches by employing a decoupled two-stage architecture: a shallow pre-training phase followed by meta-path feature aggregation with a lightweight attention module, yielding a more expressive solution for relational table learning.
\section{Preliminary}

\noindent\textbf{Definition 1} \textbf{\textit{Relational Table Data.}}
Relational table data comprises multiple structured tables linked by primary–foreign key (pkey–fkey) relations. 
It can be modeled as a heterogeneous graph $\mathcal{G} = (\mathcal{V}, \mathcal{E}, \mathcal{T}_v, \mathcal{T}_e)$, where $\mathcal{V}$ and $\mathcal{E}$ denote the sets of nodes and edges, and $\mathcal{T}_v$, $\mathcal{T}_e$ are their type sets~\cite{rdl}. 
Each table corresponds to a specific node type, and each row within the table represents one node of that type. Likewise, each primary–foreign key (pkey–fkey) relationship defines an edge type: for every such relationship, a typed edge is created between each pair of rows that are linked through the pkey–fkey match. This graph-based formulation preserves the relational structure for downstream modeling.

\vspace{0.5em}
\noindent\textbf{Definition 2} \textbf{\textit{Meta-paths.}}
A meta-path~\cite{metapath2vec} is a schema-level path defined as a sequence of edge types connecting a source node type to a target node type through intermediate types. Let $\Phi$ denote the set of all meta-paths under consideration. Each meta-path $\mathcal{P} \in \Phi$ is defined as:
\begin{equation}
    \mathcal{P}: \tau_s \rightarrow \tau_1 \rightarrow \tau_2 \rightarrow \cdots \rightarrow  \tau_t
    \label{eq:metapath}
\end{equation}
where $\tau_s$ is the source node type and $\tau_t$ is the target node type. Although the nodes connected via a meta-path may not be directly adjacent in the graph, the path implies a semantically meaningful relation between them~\cite{sehgnn}. Based on Definition~1, in the context of relational table data, the target node type corresponds to the target or task table, while the source node types refer to other auxiliary tables.

\vspace{0.5em}
\noindent\textbf{Definition 3} \textbf{\textit{Heterogeneous Graph Neural Networks.}} A heterogeneous graph neural network (HGNN) is a class of GNNs designed for representation learning over heterogeneous graphs. Let $\mathbf{h}_v \in \mathbb{R}^d$ denote the representation of node $v$ at a given layer.\footnote{For notational convenience, we use $d$ to denote feature dimension through this paper, although the actual dimensions may vary across modules.}
At each layer, an HGNN updates node representations following the message passing paradigm~\cite{msp}, where messages are aggregated from type-specific neighbor sets associated with different edge types $r \in \mathcal{T}_e$. Formally, an $l$-layer HGNN model can be written as:
\begin{equation}
\begin{aligned}
\mathrm{MP}(\mathbf{h}_v, \mathcal{G})
    &= \mathrm{UPD}\!\left(
        \left\{\mathrm{AGGR}_r\!\left(\mathbf{h}_v, \mathcal{N}_v^{\,r}\right)
            \,\middle|\,
            r \in \mathcal{T}_e
        \right\}
    \right)\\
    \mathrm{HGNN}(\mathbf{h}_v, \mathcal{G})
    &= \mathrm{MP}_l \circ \cdots \circ \mathrm{MP}_1 (\mathbf{h}_v, \mathcal{G}) \\
\end{aligned}
\label{eq:messagepassing}
\end{equation}
where $\circ$ denotes function composition, $\mathrm{MP}_i(\cdot)$ is the $i$-th message passing layer, $\mathcal{N}_v^{\,r}$ is the neighbors of node $v$ under edge type $r$, $\mathrm{AGGR}_r(\cdot)$ is an edge type specific aggregation operator (e.g. mean pooling), and $\mathrm{UPD}(\cdot)$ is an update function to integrate aggregated messages across all edge types.


\vspace{0.5em}
\noindent\textbf{Definition 4} \textbf{\textit{Attention Mechanism.}}
Attention enables dynamic weighting of input features based on pairwise relevance and serves as a core component of the Transformer architecture~\cite{transformer}. Given query matrix $Q$, key matrix $K$, and value matrix $V$, the attention output is computed as:
\begin{equation}
    \mathrm{Attn}(Q, K, V) = \mathrm{softmax}\left(\frac{QK^\top}{\sqrt{d}}\right)V.
    \label{eq:attn}
\end{equation}
This formulation supports both self-attention (when $Q = K = V$) and cross-attention (when $Q$ and $K,V$ differ), and naturally accommodates varying sequence lengths. For clarity, we omit batch dimensions and multi-head extensions, which are straightforward to incorporate.
\section{Methodology}

We propose \textbf{MetaRTL}, a scalable and expressive two-stage framework for relational table learning. The proposed framework operates in two successive stages:
(1) Pre-training Stage, in which a unified table encoder is trained with a two-layer HGNN using only a few training iterations to produce semantically meaningful and robust node embeddings; and
(2) Aggregation Stage, in which precomputed meta-path features are efficiently fused with global node representations through a compact dual-branch attention module, \textbf{MetaAttn}, to derive the final node representations for downstream tasks.

\subsection{Pre-training Stage}

As illustrated on the left side of Figure~\ref{fig:1}, MetaRTL takes as input a collection of structured relational tables, which are transformed into a heterogeneous graph. 
Given the typically large scale of such relational table data, we adopt a uniform neighbor sampling strategy to construct a series of mini-batch subgraphs 
$\mathcal{G}_i = (\mathcal{V}_i, \mathcal{E}_i, \mathcal{T}_v, \mathcal{T}_e)$ 
sampled from the entire graph $\mathcal{G}$ 
for scalable training, where corresponding table records set $\mathcal{V}_i$ and relational pkey–fkey links set $\mathcal{E}_i$ are also retrieved to provide initial input features.

To obtain node representations that are both semantically informative and stable, 
we introduce a unified table encoder. 
As illustrated in the Pre-training Stage of Figure~\ref{fig:1}, 
the encoder comprises a collection of table-specific TNN modules. 
Each $\mathrm{TNN}_\tau(\cdot)$ encodes the corresponding set of table records 
$\mathcal{V}_i^\tau$ in the sampled subgraph, with the table or node type 
$\tau \in \mathcal{T}_v$. 
The initial node embeddings $X_\tau \in \mathbb{R}^{|\mathcal{B}_\tau| \times d}$ are computed as:
\begin{equation}
    X_\tau = \mathrm{TNN}_\tau(\mathcal{V}_i^\tau), \quad \forall \tau \in \mathcal{T}_v
    \label{eq:tableencoder}
\end{equation}
where, $\mathcal{B}_\tau$ denotes the node index set of type $\tau$ contained in 
$\mathcal{V}_i$, $|\mathcal{B}_\tau|$ is the number of such nodes.

The table encoder is optimized by integrating a shallow HGNN module following the message passing paradigm defined in Definition~3. The HGNN takes as input the initial node embeddings $ \{X_\tau\}_{\tau \in \mathcal{T}_v} $ with the sampled sub graph $\mathcal{G}_i$, producing task-specific node representations:
\begin{equation}
    \mathrm{HGNN}\bigr(\{X_\tau\}_{\tau \in \mathcal{T}_v},\, \mathcal{G}_i \bigr) .
\end{equation}
These representations are passed through a prediction head, and the table encoder parameters are iteratively optimized by minimizing the task-specific loss.

Crucially, in contrast to previous RTL methods that rely on deep HGNN stacks and extensive training for full convergence, our approach employs a two-layer HGNN architecture with a quite limited number of training iterations during this stage. The objective is not to achieve complete convergence, but rather to update and refine the initial node embeddings $\{X_\tau \}_{\tau \in \mathcal{T}_v}$, producing representations that are both stable and semantically rich for downstream tasks. This lightweight pre-training paradigm balances computational efficiency with adequate representational capacity. 




\subsection{Aggregation Stage}

This section outlines the Aggregation Stage, as illustrated in Figure~\ref{fig:1}, which consists of two key components: the computation of meta-path features and the MetaAttn module. First, meta-path features are derived through a non-parametric propagation. Then, the one-layer lightweight MetaAttn module integrates these meta-path features with global node representations to generate the final embeddings for target nodes.

\subsubsection{\textbf{Meta-path Feature Computation}}
To capture high-order semantic dependencies across different auxiliary tables to target table within the heterogeneous graph constructed from relational data, we perform non-parametric propagation on the initial node embeddings to derive meta-path features. Formally, for an $l$-hop meta-path $\mathcal{P} \in \Phi$ from source node type $\tau_s$ to target node type $\tau_t$, the corresponding meta-path adjacency matrix $A^\mathcal{P}$ is obtained by multiplying the associated relation matrices in reverse order:
\begin{equation}
A^\mathcal{P} = A_{\tau_t, \tau_{l-1}} \cdots A_{\tau_2, \tau_1} A_{\tau_1, \tau_s}
\end{equation}
where $A_{\tau_l, \tau_{l-1}}$ denotes the degree-normalized adjacency matrix from node type $\tau_{l-1}$ to $\tau_l$ in the full heterogeneous graph.
This computation is inherently iterative and compositional: meta-path adjacency matrices can be efficiently constructed by sequentially multiplying the intermediate products of shorter meta-path segments. By recursively applying this procedure, we obtain the complete set of meta-path adjacency matrices $\{A^\mathcal{P}\}_{\mathcal{P} \in \Phi}$ for all meta-paths up to a predefined maximum length. This step is independent of subgraph sampling and model training, and can be efficiently performed during data preprocessing over the entire heterogeneous graph.

For each mini-batch subgraph generated by neighbor sampling, the precomputed meta-path adjacency matrices $A^\mathcal{P}$ are sliced according to the indices of the source and target node types present in the current subgraph, denoted by $\mathcal{B}_{\tau_s}$ and $\mathcal{B}_{\tau_t}$, respectively. The sliced adjacency matrices are then multiplied by the initial embeddings $X_{\tau_s}$ of the meta-path source node type $\tau_s \in \mathcal{T}_v$ to yield meta-path feature $H^\mathcal{P} \in \mathbb{R}^{|\mathcal{B}_{\tau_t}| \times d}$:
\begin{align}
H^\mathcal{P}
= A^\mathcal{P}[\mathcal{B}_{\tau_t}, \mathcal{B}_{\tau_s}] X_{\tau_s}
\label{eq:slice}
\end{align}
where $[,]$ denotes the slicing operation.
The resulting meta-path feature set $\{H^\mathcal{P}\}_{\mathcal{P} \in \Phi}$ effectively encodes the semantic information carried from auxiliary tables $\tau_s$ to the target table $\tau_t$ along all meta-paths within the sampled subgraph $\mathcal{G}_i$.

\begin{figure}[tp]
    \centering
    \includegraphics[width=0.8\linewidth]{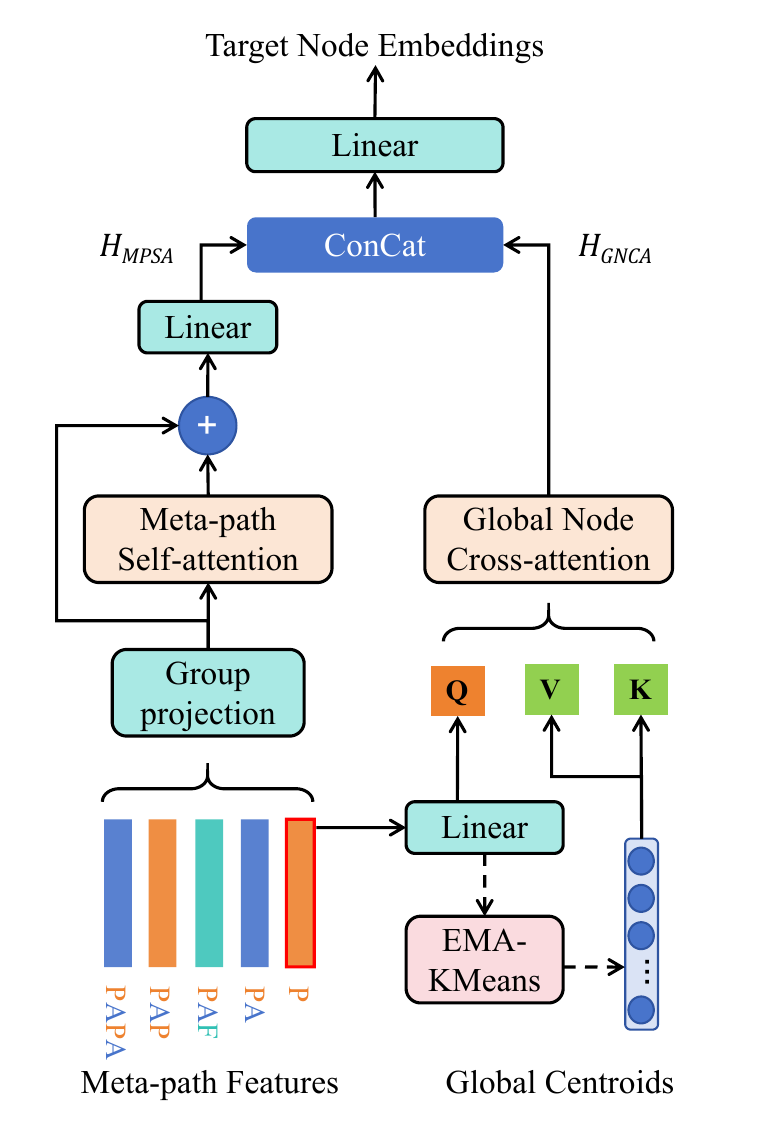}
    \caption{Architecture of the MetaAttn Module.}
    \label{fig:model}
\end{figure}

\subsubsection{\textbf{MetaAttn Module}}
To effectively aggregate the meta-path features of target table corresponding nodes while incorporating global semantic information, we design a lightweight, single-layer attention-based module named MetaAttn. As illustrated in Figure~\ref{fig:model}, it consists of a meta-path self-attention module that captures the relative semantic importance among multiple meta-path features, a global node cross-attention module that incorporates full-graph semantic context, and a feature aggregation step that combines the outputs to form the final target node representations.

\vspace{0.5em}
\noindent\textit{Meta-path Self-attention (MPSA).}
\vspace{0.1em}

Motivated by the observation that auxiliary tables contribute unequally to relational table learning, we introduce a meta-path self-attention mechanism along the meta-path axis to selectively aggregate semantic features from different auxiliary tables into the target table. This mechanism highlights task relevant information while suppressing noise from less informative tables, as illustrated in the left part of Figure~\ref{fig:model}.

Given the precomputed meta-path features $\{H^{\mathcal{P}}\}_{\mathcal{P} \in \Phi}$ from Equation~\ref{eq:slice}, we utilize the mapping that associates each meta-path $\mathcal{P}$ with its corresponding source node type $\tau_s$. Based on this association, we group the meta-path features by their source types and apply type-specific linear transformations to each group.
The transformed representations are then concatenated along a new axis to construct a aggregated meta-path feature tensor $H_{\text{meta}} \in \mathbb{R}^{|\mathcal{B}_{\tau_t}| \times |\Phi| \times d}$:
\begin{equation}
H_{\text{meta}}
= \Big\|_{\tau_s \in \mathcal{T}_v} 
\mathrm{Linear}_{\tau_s}\left( \{H^\mathcal{P} \mid \mathcal{P} \mapsto \tau_s \} \right)
\label{eq:meta-concat}
\end{equation}
where $\mathcal{P} \mapsto \tau_s$ denotes the mapping that indicates each meta-path $\mathcal{P}$ to its source node type $\tau_s$, $\mathrm{Linear}_{\tau_s}(\cdot)$ denotes the linear layer associated with type $\tau_s$, and $\big\|$ represents the concatenation operation. Then to capture the relative importance of different auxiliary tables, a self-attention mechanism formulated in Definition~4 is applied over the meta-path dimension on the meta-path feature tensor computed above, followed by a residual connection and a linear projection. The resulting meta-path semantic aggregation representation $H_{\text{MPSA}} \in \mathbb{R}^{|\mathcal{B}_{\tau_t}| \times d}$ is computed as:
\begin{equation}
H_{\text{MPSA}} = \mathrm{Linear}\Big(\mathrm{SelfAttn}(H_{\text{meta}}) + H_{\text{meta}}\Big)
\label{eq:mpsa}
\end{equation}
where, the attention output $H_{\text{MPSA}}$ selectively encodes the aggregated semantic information from all auxiliary tables for target table, i.e. each target node.

\vspace{0.5em}
\noindent\textit{Global Node Cross-attention (GNCA).}
\vspace{0.1em}
To alleviate potential information loss introduced by neighbor sampling, 
we adopt a global node cross-attention mechanism inspired by the 
attention module in previous work~\cite{goat, graphtransformerslargegraphs}, 
as shown on the right side of Figure~\ref{fig:model}. 
This mechanism enables each target node in a sampled subgraph to attend to 
a set of shared global centroids $\mathcal{C}$, which compactly capture the overarching 
semantic structure of the entire heterogeneous graph. 
The centroids are updated by clustering the current target-node features during training, 
thereby providing a holistic semantic context that complements the local neighborhood information.

For the 0-hop meta-path, the feature $H^\mathcal{P}_{\tau_t} = X_{\tau_t}$ reduces to 
the original node features of the target table and is directly used as the input.  
The resulting global semantic representation, denoted as 
$H_{\text{GNCA}} \in \mathbb{R}^{|\mathcal{B}_{\tau_t}| \times d}$, is computed as:
\begin{equation}
\begin{aligned}
H_{\text{GNCA}} &= \mathrm{CrossAttn}\left(\mathrm{Linear}(H^\mathcal{P}_{\tau_t}),\, \mathcal{C}\right) \\
\mathcal{C} &\leftarrow \mathrm{KMeans}\big(H^\mathcal{P}_{\tau_t},\, \mathcal{C}\big)
\end{aligned}
\label{eq:gnca}
\end{equation}
where $\mathrm{KMeans}(\cdot)$ denotes the centroid-update rule following 
the exponential moving average strategy, 
and $\mathrm{CrossAttn}(\cdot)$ denotes the cross-attention operation in which 
the linearly transformed node features serve as the query matrix, 
while the global centroids $\mathcal{C}$ (stacked along the first dimension) 
are used as both key and value matrices.

\vspace{0.5em}  
\noindent\textit{Feature Aggregation.}
\vspace{0.1em}
Finally, the meta-path self-attention semantic feature $H_{\text{MPSA}}$ obtained from Equation~\ref{eq:mpsa} and the global node cross-attention semantic feature $H_{\text{GNCA}}$ obtained from Equation~\ref{eq:gnca} are concatenated and aggregated through a linear layer to produce the target node embeddings output, formulated as $\mathrm{Linear}([ H_{\text{MPSA}} \parallel H_{\text{GNCA}} ])$.

Depending on the downstream task, such as regression or classification, different output layers are applied to generate the final predictions. The model is trained end-to-end by minimizing the corresponding loss via backpropagation.
\section{Experiments}

\subsection{Experimental Setup}

\subsubsection{\textbf{Datasets}}
We evaluate MetaRTL on two public relational table learning benchmarks: SJTUTables~\cite{rllm} and RelBench~\cite{rdl}.  
SJTUTables provides three small to medium scale multi-class classification datasets: \texttt{TACM12k}, \texttt{TLF2K}, and \texttt{TML1M}. 
RelBench consists of 7 large-scale datasets covering 12 binary classification and 9 regression tasks across real-world domains such as e-commerce, social networks, and healthcare. During training, we apply temporal uniform neighbor sampling, and all data preprocessing, sampling strategies, and evaluation metrics strictly adhere to the official benchmark settings.  
The detailed statistics of the datasets in both benchmarks are summarized in Table~\ref{tab:datasets}.

\begin{table}[t]
\caption{Statistical analysis of two benchmark datasets}
\centering
\begin{tabular}{lcrcc}
\toprule
Dataset & \#Tables & \#Rows & \#Columns & Max hop\\
\midrule
TACM12k             & 4         & 97,774         &  12   & 2 \\
TLF2K               & 3         & 101,773        &  15   & 1 \\
TML1M               & 3         & 1,010,132      &  20   & 2 \\
\midrule
rel-f1              & 9         & 97,606         &  77   & 3 \\
rel-trial           & 15        & 5,852,157      & 	140  & 3 \\
rel-avito           & 8         & 20,679,117     &  43   & 2 \\
rel-amazon          & 3         & 24,291,489     &  15   & 2 \\
rel-hm              & 3         & 33,265,846     &  37   & 2 \\
rel-stack           & 7         & 38,109,828     &  51   & 3 \\
rel-event           & 5         & 41,328,337     & 	128  & 2  \\
\bottomrule
\end{tabular}
\label{tab:datasets}
\end{table}

\subsubsection{\textbf{Baselines}}
We compare MetaRTL with a diverse set of representative baselines, which can be broadly categorized into single-table and multi-table methods. 
\textbf{Single-table methods} operate only on the target table, including LightGBM~\cite{lightgbm}, a gradient boosting baseline, FTTransformer~\cite{fttransformer}, which applies row-wise Transformers over tokenized features, Trompt~\cite{trompt}, a prompt-based tabular model, and ExcelFormer~\cite{excelformer}, which enhances column selection and feature interactions via attention. 
\textbf{Multi-table methods} exploit auxiliary tables via pkey-fkey relations, including RDL~\cite{rdl}, combining ResNet-based TNN with heterogeneous GraphSAGE, BRIDGE~\cite{rllm}, using a single table with standard GCN, RelGNN~\cite{relgnn}, with composite relational message passing, and LightRDL~\cite{lightrdl}, which incorporates temporal relational graphs and heterogeneous GraphSAGE with LightGBM derived embeddings.

For single-table methods, following the original RelBench setup, since the task tables contain only indices and labels without feature columns, we perform a left join operation to merge the feature tables with the target tables.
Since the official implementations of RelGNN and LightRDL are incomplete, we reproduce them based on the released code and publications.
BRIDGE cannot generalize to relational data involving more than two entity tables, and thus, in our experiments on RelBench, we use only the auxiliary table that is most relevant to the target table.
In addition, LightRDL requires timestamped datasets for proper data splitting. As some datasets in SJTUTables lack temporal features, we substitute them with randomly generated timestamps.
The main experimental pipeline is implemented based on the open-source relational table learning library relationLLM (rLLM)\footnote{\url{https://github.com/rllm-project/rllm}}.

\subsection{Performance Evaluation on Benchmarks}

\subsubsection{\textbf{Results on SJTUTables}}
\begin{table}[t]
\centering
\caption{Performance on SJTUTables (Accuracy\% ↑)}
\label{tab:sjtutables}
\begin{tabular}{lcccc}
\toprule
Group  &  Method & TACM12k & TLF2K & TML1M \\
\midrule
\multirow{4}{*}{Single-table} 
    & LightGBM              & 35.70 & 35.52 & 26.66 \\
    & FTTransformer         & 14.88 & 15.80 & 28.62 \\
    & Trompt                & 12.42 & 13.41 & 27.85 \\
    & ExcelFormer           & 17.13 & 20.61 & 31.90 \\
\midrule
\multirow{4}{*}{Multi-table} 
& RDL        & 19.23 & 19.58 & 26.93 \\
& RelGNN     & 18.19 & 20.31 & 27.12 \\
& BRIDGE     & 25.60 & 42.20 & 36.20 \\
& LightRDL   & 24.94 & 39.17 & 35.77 \\
& MetaRTL    & \textbf{46.54} & \textbf{43.17} & \textbf{37.09} \\
\bottomrule
\end{tabular}
\end{table}

Table~\ref{tab:sjtutables} presents the results on the three datasets from the SJTUTables benchmark. 
We can clearly see that MetaRTL consistently achieves the best performance across all datasets, significantly outperforming the baselines. On \texttt{TACM12k}, the absolute accuracy of MetaRTL improves over LightGBM by nearly 11\% and surpasses the strongest deep learning baseline BRIDGE by approximately 21\%. It also demonstrates superior generalization on \texttt{TLF2K} and \texttt{TML1M}. This improvement is primarily due to MetaRTL's design: by leveraging the rich features during pretraining, it obtains stable initial node semantic embeddings, and its meta-path based lightweight aggregation prevents over-smoothing of neighbor information, thereby preserving feature signals and enhancing predictive performance.

We further observe that RDL and RelGNN underperform on SJTUTables. This can be primarily attributed to a mismatch between their complex model architectures and the simpler patterns in the benchmark. SJTUTables contains a limited number of tables with relatively shallow and sparse relations, under which the deep HGNN backbones in RDL and RelGNN may introduce excessive modeling capacity, resulting in overfitting.
In contrast, MetaRTL better leverages these compact but informative representations through non-parametric propagation and meta-path-aware aggregation.

\subsubsection{\textbf{Results on RelBench}}
\begin{table*}[t]
\centering
\caption{Results on RelBench classification tasks (ROC-AUC ↑)}
\label{tab:relbenchcls}

\begin{tabular}{llccccccccc}
\toprule
\multirow{2}{*}{Dataset} & \multirow{2}{*}{Task} 
& \multicolumn{4}{c}{Single-table} & \multicolumn{4}{c}{Multi-table} \\ 
\cmidrule(lr){3-6} \cmidrule(lr){7-11} 
 &  & LightGBM & FTTransformer & Trompt & ExcelFormer           & RDL & RelGNN    & BRIDGE & LightRDL & MetaRTL \\
\midrule
\multirow{2}{*}{rel-f1} 
    & driver-top3     & 0.7392 & 0.7362 & 0.7317 & 0.7522       & 0.7723  & 0.8086  & 0.7713  & 0.7916  & \textbf{0.8415} \\
    & driver-dnf     & 0.6856 & 0.6831 & 0.6752 & 0.6720        & 0.7125  & 0.6706 & 0.7016 & 0.6863 & \textbf{0.7418} \\
\midrule
\multirow{1}{*}{rel-trial} 
    & study-outcome     & 0.7009 & 0.6239 & 0.6573 & 0.6731     & 0.6902 & 0.6916  & 0.6898 & 0.6815  & \textbf{0.7013} \\
\midrule
\multirow{2}{*}{rel-avito} 
    & user-clicks     & 0.5360 & 0.5132 & 0.5423 & 0.5532       & 0.6576  & 0.6613 & 0.5501 & 0.6411  & \textbf{0.6614} \\
    & user-visits     & 0.5305 & 0.5210 & 0.5611 & 0.5501       & \textbf{0.6599} & 0.6547 & 0.5616 & 0.6239 & 0.6587 \\
\midrule
\multirow{2}{*}{rel-amazon} 
    & user-churn     & 0.5222 & 0.5230 & 0.5310 & 0.5209        & 0.7042 & 0.6544 & 0.5351 & 0.6917 & \textbf{0.7101} \\
    & item-churn     & 0.6254 & 0.6415 & 0.5982 & 0.6250        & 0.8281 & 0.8264 & 0.6993 & 0.8120 & \textbf{0.8290} \\
\midrule
\multirow{1}{*}{rel-hm} 
    & user-churn     & 0.5521 & 0.5013 & 0.5419 & 0.5395        & \textbf{0.7011} & 0.6810 & 0.5933 & 0.6713 & 0.6795 \\
\midrule
\multirow{2}{*}{rel-stack} 
    & user-engagement   & 0.6339 & 0.6002 & 0.6573 & 0.6920     & 0.9047 & 0.8995  & 0.7974 & 0.8902 & \textbf{0.9120} \\
    & user-badge        & 0.6343 & 0.6036 & 0.6282 & 0.6713     & 0.8886  & 0.8898 & 0.8790 & 0.8671 & \textbf{0.8898} \\
\midrule
\multirow{2}{*}{rel-event} 
    & user-repeat     & 0.6804 & 0.6910 & 0.6717 & 0.6973       & 0.7560  & 0.7307 & 0.7530  & 0.7132 & \textbf{0.7604} \\
    & user-ignore     & 0.7993 & 0.7569 & 0.7731 & 0.7792       & 0.7514  & 0.7911 & 0.7616 & 0.7630 & \textbf{0.8703} \\
\midrule
\multicolumn{2}{l}{\textbf{Average ranking}}   &6.25      & 8.00      &7.33    &6.83     &2.92        &3.33       &4.83       &4.17       & \textbf{1.33}\\
\bottomrule
\end{tabular}
\end{table*}

\begin{table*}[t]
\centering
\caption{Results on RelBench regression tasks (MAE ↓)}
\label{tab:relbenchreg}

\begin{tabular}{llccccccccc}
\toprule
\multirow{2}{*}{Dataset} & \multirow{2}{*}{Task} 
& \multicolumn{4}{c}{Single-table} & \multicolumn{4}{c}{Multi-table} \\ 
\cmidrule(lr){3-6} \cmidrule(lr){7-11} 
 &  & LightGBM & FTTransformer & Trompt & ExcelFormer          & RDL & RelGNN & BRIDGE & LightRDL & MetaRTL \\
\midrule
\multirow{1}{*}{rel-f1} 
    & driver-position     & 4.170 & 4.210 & 4.191 & 4.112      & 4.172 & 4.250  & 4.179 & 4.170 & \textbf{3.987} \\
\midrule
\multirow{2}{*}{rel-trial} 
    & study-adv1erse     & 44.011 & 44.490 & 45.397 & 45.552     & 44.473 & 44.461 & 44.773 & 43.910 & \textbf{43.102} \\
    & site-success     & 0.425   & 0.431 & 0.440   & 0.429      & 0.400  & 0.355  & 0.431   & 0.411 & \textbf{0.350} \\
\midrule
\multirow{1}{*}{rel-avito} 
    & ad-ctr          & 0.041    & 0.049 & 0.045 & 0.050        & 0.041 & \textbf{0.039} & 0.044 & 0.040 & 0.040 \\
\midrule
\multirow{2}{*}{rel-amazon} 
    & user-ltv     & 16.783     & 16.930 & 16.606 & 17.001       & 14.313 & 16.783 & 15.374 & 14.310 & \textbf{14.102} \\
    & item-ltv     & 60.569     & 60.319 & 61.424 & 59.983       & 50.053 & 48.826 & 60.441 & 48.112 & \textbf{46.176} \\
\midrule
\multirow{1}{*}{rel-hm} 
    & item-sales     & 0.076    & 0.092 & 0.073 & 0.073         & 0.056 & 0.054 & 0.071 & 0.044 & \textbf{0.039} \\
\midrule
\multirow{1}{*}{rel-stack} 
    & post-votes   & 0.068      & 0.074 & 0.080  & 0.070         & 0.065 & 0.065 & 0.066 & 0.064 & \textbf{0.060} \\
\midrule
\multirow{1}{*}{rel-event} 
    & user-attendance   & 0.264  & 0.270 & 0.281 & 0.264        & 0.258 & 0.248 & 0.259 & 0.260 & \textbf{0.241} \\
\midrule
\multicolumn{2}{l}{\textbf{Average ranking}}  &5.72   & 7.72  &7.72  &6.78     &3.89  &3.78  &5.50  &2.61  & \textbf{1.67}\\
\bottomrule
\end{tabular}
\end{table*}

Tables~\ref{tab:relbenchcls} and \ref{tab:relbenchreg} present the detailed results of classification and regression tasks, respectively, across 21 tasks from the RelBench benchmark. The tables report ROC-AUC for classification tasks, MAE for regression tasks, and the average ranking of each method, providing a comprehensive comparison of baseline approaches and our proposed MetaRTL. MetaRTL consistently achieves the lowest average ranks on both classification and regression tasks, and attains the best performance on 18 out of 21 tasks. For the remaining tasks, MetaRTL still produces results that are close to the top-performing method. Overall, these results highlight not only the effectiveness of MetaRTL in individual tasks but also its strong generalization capability across diverse relational table datasets, demonstrating its robustness in modeling complex real-world relational structures.

On the RelBench benchmark, which better reflects real-world scenarios, single-table methods show clear differentiation. When individual tables contain sufficient information (e.g., \texttt{rel-event}), they can achieve performance comparable to some multi-table methods. However, on datasets with less informative tables and richer textual features (e.g., \texttt{rel-stack} and \texttt{rel-amazon}), their performance drops significantly. This underscores the importance of leveraging pkey–fkey based multi-table relationships. In contrast, MetaRTL demonstrates strong performance by effectively extracting informative signals from auxiliary tables while suppressing noise. It achieves notable gains on datasets with complex multi-table structures (e.g., \texttt{rel-f1}), while improvements are smaller on simpler datasets (e.g., \texttt{rel-hm}), where it may even slightly under-perform some baselines. Overall, these results highlight the effectiveness of meta-path-based aggregation for modeling complex inter-table relationships.


\subsection{Ablation of Model Components}
\subsubsection{\textbf{Impact of Model Components}}

\begin{table}[!t]
\centering
\caption{Results of ablation study}
\label{tab:abl}
\begin{tabular}{lccc}
\toprule
Variant & driver-top3$\uparrow$ & driver-dnf$\uparrow$ & driver-position$\downarrow$ \\
\midrule
w/o TNN Pretrain            & 0.7311 & 0.6761 & 4.319 \\
w/o Meta-path Features      & 0.6294 & 0.6013 & 4.870 \\
w/o Global Attention        & 0.8389 & 0.7216 & 4.110 \\
\midrule
MetaRTL              & \textbf{0.8415} & \textbf{0.7418} & \textbf{3.987} \\
\bottomrule
\end{tabular}
\end{table}

To quantify the contribution of each module in MetaRTL, we perform ablation studies on the \texttt{rel-f1} dataset. Similar trends are observed on other datasets.
Three variants are considered to evaluate the core components: (i) without TNN pretraining (w/o TNN Pretrain), (ii) without meta-path-based features (w/o Meta-path Features), and (iii) without the global node cross-attention mechanism (w/o Global Attention). As shown in Table~\ref{tab:abl}, removing either TNN pretraining or meta-path features leads to a substantial drop in performance, highlighting the importance of stable node initialization and structural semantics. Omitting the global attention module also slightly degrades performance, reflecting its complementary role in aggregating global context beyond local neighborhoods. 
Overall, these results demonstrate that each component of MetaRTL contributes meaningfully to the model’s effectiveness.

\subsubsection{\textbf{Effect of Pretraining and Aggregation Stages}}
\begin{figure}[tp]
    \centering
    \includegraphics[width=0.9\linewidth]{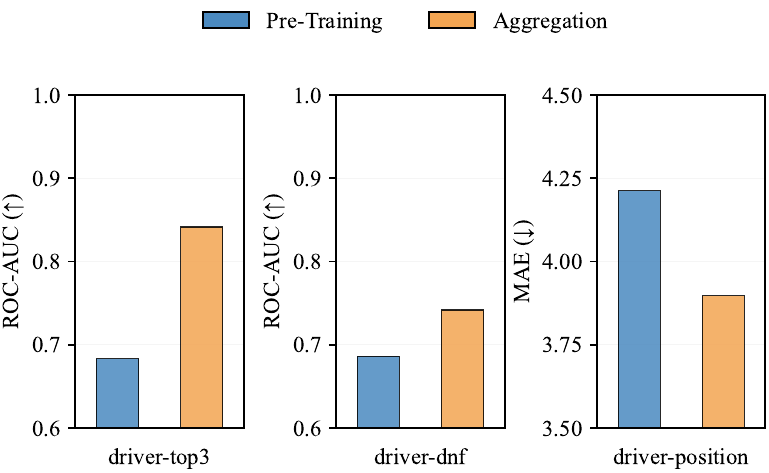}
    \caption{Prediction Performance: Pre-training Stage vs. Aggregation Stage.}
    \label{fig:exp3}
\end{figure}
Further, to verify that the performance gains of MetaRTL primarily stem from the Aggregation Stage rather than the initial Pre-training Stage, we compare their prediction results in Figure~\ref{fig:exp3}. Across all three tasks, the Pre-training Stage yields lower performance than the second stage. This shows that the Pre-training Stage, with limited training iterations and a shallow HGNN, mainly produces stable initial node features containing basic semantic information, rather than a fully converged, final semantic representation. Building on this, the Aggregation Stage leverages precomputed meta-path features with meta-path self-attention to integrate local heterogeneous semantics effectively. The global node cross-attention further mitigates biases introduced by neighbor sampling. Together, these mechanisms significantly boost model performance. In summary, MetaRTL’s two-stage design distinctly decouples semantic stabilization from deep semantic fusion, resulting in superior predictive accuracy.

\subsection{Training Efficiency and Scalability Analysis}

\subsubsection{\textbf{Effect of Pretraining Duration}}
\begin{figure}[t]
    \centering
    \includegraphics[width=0.8\linewidth]{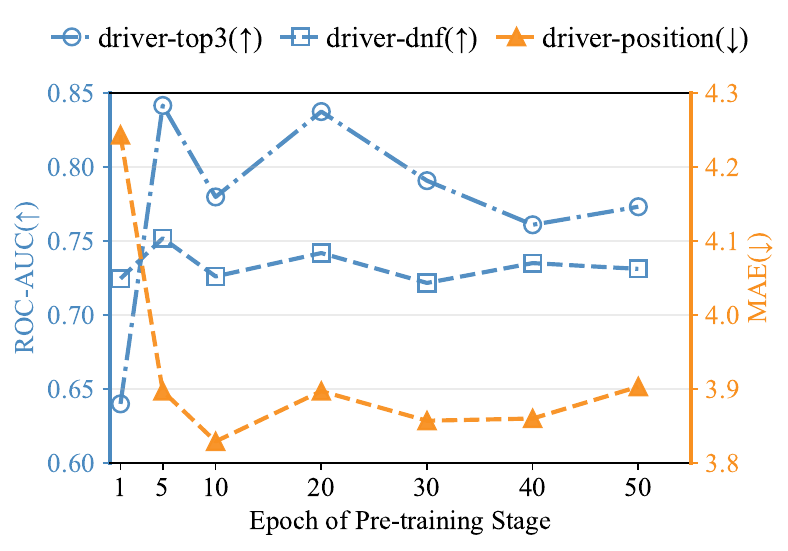}
    \caption{MetaRTL Performance vs. Pre-training Stage Epochs.}
    \label{fig:k}
\end{figure}

To systematically evaluate the efficiency of the Pre-training Stage, we assess variants on three downstream tasks from the \texttt{rel-f1} dataset. As shown in Figure~\ref{fig:k}, when pre-training is too short particularly at 1 epoch, the table encoder is underfitted and fails to capture sufficient structural and semantic information. 
This results in low quality initial node embeddings for the Aggregation Stage and significantly degraded downstream performance. Extending pre-training to 5–10 epochs substantially improves and stabilizes performance, indicating that a moderate number of iterations allows the encoder to effectively extract essential patterns from tabular inputs. However, further increasing pre-training leads to a gradually slight performance decline, which can be attributed to overfitting: excessive training causes the encoder to memorize spurious patterns and introduce strong inductive biases that cannot be easily corrected during downstream learning. In summary, the number of epochs has a critical impact on MetaRTL, with too few causing underfitting and too many leading to harmful bias.

\begin{figure}[tp]
    \centering
    \includegraphics[width=0.8\linewidth]{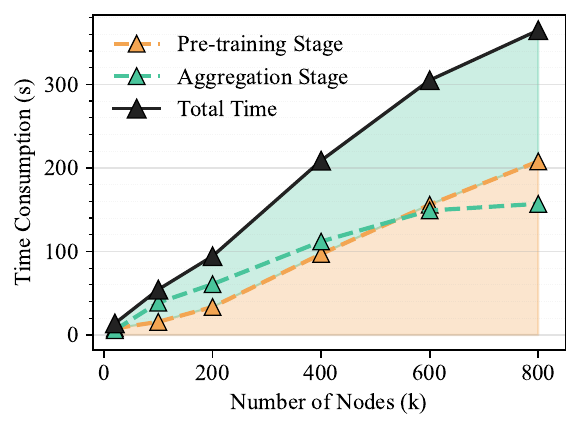}
    \caption{Training Time vs. Number of Nodes (Fixed Node Out-Degree).}
    \label{fig:time}
\end{figure}

\subsubsection{\textbf{Scalability Test}}
To experimentally validate the scalability of our method, we generated synthetic graphs with a fixed average out-degree, varying the total number of nodes, with target nodes accounting for 1\% of the total. 
As shown in Figure~\ref{fig:time}, the overall training time of MetaRTL increases approximately linearly with the number of nodes.
Specifically, the Pre-training Stage initially consumes a small fraction of the total time but grows faster as the graph size increases, while the Aggregation Stage remains relatively lightweight. Non-parametric meta-path computation contributes negligibly to the overall time. 
These results indicate that MetaRTL scales efficiently across graphs of varying sizes, even under variations in table features and their distributions.

\section{Conclusion}

We present MetaRTL, a novel framework for relational table learning that effectively models complex relational structures induced from relational table data. Instead of relying on deep, parameter-heavy message passing, MetaRTL adopts a two-stage design: it first learns stable node representations through shallow pretraining, and then aggregates precomputed meta-path features using a lightweight attention module, MetaAttn. This decoupled architecture enables expressive relational modeling while preserving scalability across large and diverse relational databases. Extensive experiments demonstrate that MetaRTL consistently outperforms existing methods, establishing new state-of-the-art results in relational table learning.
As future work, we plan to extend this framework to data lake scenarios~\cite{pan2026lakemlb}.

\section*{Acknowledgments}
This work was supported by the Natural Science Foundation of Shanghai under Grant 23ZR1434000.

\bibliographystyle{IEEEtran}
\bibliography{ref}

\end{document}